\documentclass[conference]{IEEEtran}
\usepackage{cite}
\usepackage{amsmath,amssymb,amsfonts}
\usepackage{algorithmic}
\usepackage{graphicx}
\usepackage{textcomp}
\usepackage{xcolor}
\usepackage{comment}
\usepackage{array}
\usepackage{tikz}
\usepackage{subfigure}
\usetikzlibrary{positioning}
\usepackage{subfigure}
\usepackage{blindtext}

\def\BibTeX{{\rm B\kern-.05em{\sc i\kern-.025em b}\kern-.08em
    T\kern-.1667em\lower.7ex\hbox{E}\kern-.125emX}}

\begin{document}

\title{Perceptual Evaluation of Adversarial Attacks for CNN-based Image Classification
}
\author{Sid Ahmed Fezza$^{\S}$, Yassine Bakhti$^{\S \ast}$, Wassim Hamidouche$^{\ast}$ and Olivier D\'eforges$^{\ast}$\\
$^{\S}$National Institute of Telecommunications and ICT, Oran, Algeria\\
$^{\ast}$Univ. Rennes, INSA Rennes, CNRS, IETR - UMR 6164, Rennes, France\\    		
sfezza@inttic.dz, firstname.lastname@insa-rennes.fr}
\maketitle
\begin{abstract}
Deep neural networks (DNNs) have recently achieved state-of-the-art performance and provide significant progress in many machine learning tasks, such as image classification, speech processing, natural language processing, etc. However, recent studies have shown that DNNs are vulnerable to adversarial attacks. For instance, in the image classification domain, adding small imperceptible perturbations to the input image is sufficient to fool the DNN and to cause misclassification. The perturbed image, called \textit{adversarial example}, should be visually as close as possible to the original image. However, all the works proposed in the literature for generating adversarial examples have used the $L_{p}$ norms ($L_{0}$, $L_{2}$ and $L_{\infty}$) as distance metrics to quantify the similarity between the original image and the adversarial example. Nonetheless, the $L_{p}$ norms do not correlate with human judgment, making them not suitable to reliably assess the perceptual similarity/fidelity of adversarial examples. In this paper, we present a database for visual fidelity assessment of adversarial examples. We describe the creation of the database and evaluate the performance of  fifteen state-of-the-art full-reference (FR) image fidelity assessment metrics that could substitute $L_{p}$ norms. The database as well as subjective scores are publicly available to help designing new metrics for adversarial examples and to facilitate future research works.\footnote{https://github.com/safezza/IQA-CNN-Adversarial-Attacks}
\end{abstract}

\section{Introduction}
\label{sec1}
One can only be impressed by the deep neural networks (DNNs) performance that are significantly superior to those achieved using conventional shallower networks. Taking advantage of the proliferation of large datasets in addition to the increase in computational power, the DNNs have shown a high efficiency in various difficult tasks such image \hbox{classification \cite{B1}}, object detection \cite{B2}, speech recognition \cite{B3} and natural language processing \cite{B4}. For instance, in the field of image recognition, the DNNs are able to recognize images with almost human precision, allowing them to be used in different sensitive applications such as autonomous cars, biometric, video surveillance, etc.

Despite state-of-the-art performance achieved by DNNs, it has been shown that they are vulnerable and unstable to adversarial attacks \cite{B5}. For instance, in the field of image classification, Szegedy \textit{et al.} \cite{B6} was the first to show that small and almost imperceptible perturbations added to test images could lead DNN to misclassifying them. The perturbed images are called \textit{adversarial examples}.
\begin{figure}[t!]
\centering
\includegraphics[scale=0.27]{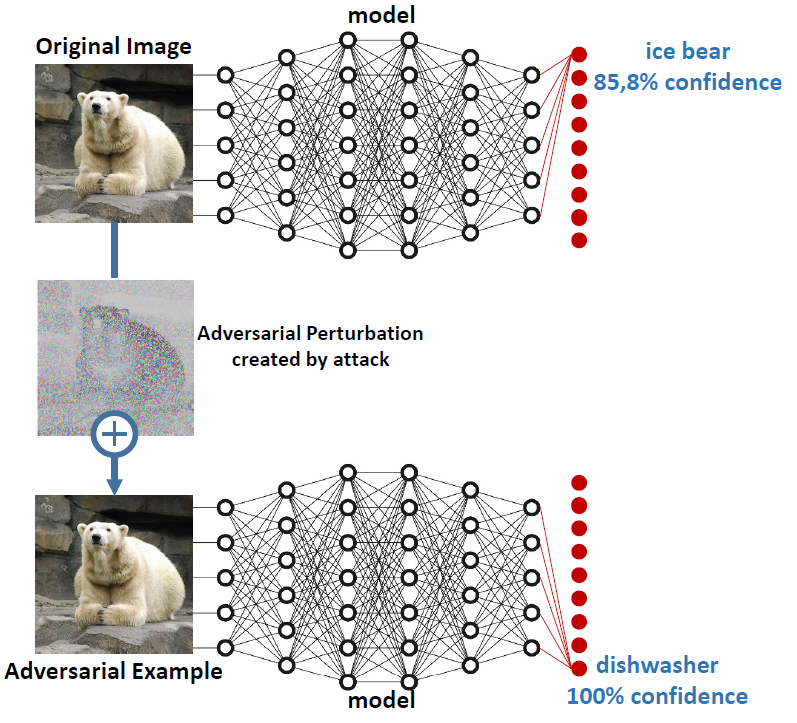}
\caption{Applying small imperceptible perturbation to the input image fool the deep neural network classifier. The original image is classified as an \textit{ice bear} with 85.8\% confidence, while the adversarial example is classified as a \textit{dishwasher} with 100\% confidence.}
\label{S1}
\end{figure}

Goodfellow \textit{et al.} \cite{B7} define adversarial examples as ``inputs to machine learning models that an attacker has intentionally designed to cause the model to make a mistake.'' Figure \ref{S1} shows how an original image carefully crafted using a small perturbation induces the network into misclassification with high confidence. Although, to a human, the adversarial image is indistinguishable from the original, \textit{i.e.} the perturbation is quasi-imperceptible, the classifier labels them differently. This highlights the lack of robustness of the DNNs against adversarial examples, which raises security issues and limits the applications in which the neural networks can be deployed in a real-world environment. For instance, an adversary can use adversarial examples to manipulate the traffic signs so that the car takes undesirable and inappropriate actions, which is significantly dangerous. Therefore, it is of paramount importance to understand how and why these vulnerabilities to attacks occurs, thereby increasing the robustness of DNNs against the adversarial examples and bridging the gap between human perception and DNN-based systems.

Recently, great efforts have been made to propose methods for generating adversarial examples, which have been used as a benchmark for evaluating the robustness of candidate defenses. Several adversarial attacks strategies have been proposed in the literature, and they are primarily differentiated by their computational cost, the level of knowledge  about the attacked model and the purpose of the attacker \cite{B8,B9,B10}.

As a general rule, two factors of adversarial examples, in the domain of image classification, should be considered \cite{B8,B9,B10}:
\begin{enumerate}
	\item  Fooling the image classifier: the adversarial example is crafted by adding small perturbation to the original image in such a way to cause classification mistake, \textit{i.e.,} the perturbed image is misclassified to a specific class (\textit{targeted attack}) or only misclassified to an arbitrary class (\textit{untargeted attack}). 
	\item  Imperceptible perturbation: the introduced perturbation should be undetectable by human observer. The original image and its intentionally perturbed version (adversarial example) are expected to be visually very close, and the differences between them are hardly noticeable by the human eye. 
\end{enumerate}

For the second aspect, which is the focus of our study, all the works proposed in the literature for generating adversarial examples have used the $L_{p}$ norms ($L_{0}$, $L_{2}$ and $L_{\infty}$) as distance metrics to quantify the similarity between the original image and the adversarial example \cite{B9,B10}. However, the $L_{p}$ norms do not correlate with human judgment, because they are pixel-based error measures and do not take into account the properties of human visual system (HVS) \cite{B11}. Despite these common measures provide poor performance for assessing perceptual similarity/fidelity, all existing works have adopted these metrics as perturbation measures for generating adversarial examples.

On the other hand, in the last decade, considerable research efforts have been made to develop objective quality/fidelity assessment metrics \cite{B12,B13,B14}. The purpose of this research is to develop tools allowing to evaluate the quality/fidelity in a way that is consistent with human judgments \cite{B14}. There is a tendency to confuse image quality metrics with image fidelity metrics, despite the fact that they are closely linked, the two families of metrics have different purposes. The former are designed to predict subjective human appreciation upon the quality of multimedia content, while the latter refer to the ability to quantify the visual differences between a reference and test image \cite{B15}. Given the purpose of this work, image fidelity assessment (IFA) metrics are more appropriate for the generation and performance analysis of adversarial examples. That is why in the rest of this paper, we refer to image fidelity metrics instead image quality metrics.

However, the use of an inappropriate IFA metric can lead to wrong conclusions and suboptimal results, which can be the case with $L_{p}$ norms that exhibit poor correlation with the human perception. There is therefore an urgent need to find a more accurate IFA metric that could substitute $L_{p}$ norms for generating and assessing adversarial examples in close agreement with human similarity judgments.

The natural way to reach this goal is to take advantage of the many IFA metrics proposed in the literature. However, these metrics were typically developed for some specific applications, and consequently were designed to capture distortions that are related to these applications, such as blur and blocking for compression, noise for acquisition and fast fading for wireless transmission, to cite a few examples. Nevertheless, the adversarial perturbations/distortions used against DNNs can have different properties than those widely tackled by the quality/fidelity assessment community. Thus, developing new reliable IFA metrics specifically for adversarial examples represents a new research challenges to this community.

In this paper, we present a database for visual fidelity assessment of adversarial examples. To the best of our knowledge, this database is the first one specifically dedicated to the perceptual assessment of adversarial perturbations against DNNs and is publicly available to facilitate future research works. The dataset includes 360 images that have been generated using six prominent adversarial attacks with different levels of perturbations. The subjective data of eighteen human subjects have been collected, where each subject was asked to rate the fidelity of the adversarial example with respect to the reference image. The resulting MOS scores have been used to evaluate the performance of the three distance metrics ($L_{0}$, $L_{2}$ and $L_{\infty}$) and to assess the performance of fifteen state-of-the-art full-reference (FR) image fidelity assessment metrics, as well as can be used to design new IFA metrics for adversarial examples.

The rest of this paper is organized as follows. \hbox{Section \ref{sec2}} provides the taxonomy of adversarial attacks. Section \ref{sec3} describes the performed subjective  experiment, including the preparation of the test material, environmental setup and the test methodology. Next, the results and analysis of objective metrics are provided in Section \ref{sec4}. Finally, Section \ref{sec5} concludes the paper.
\section{Adversarial Attacks on Deep Neural Networks}
\label{sec2}
An adversarial example is an original image carefully-crafted by an adversary attack with the aim to fool DNN classifier. The adversary attacks can be divided into two categories: \textit{white-box attacks} that have a full access to the architecture and model’s parameters of the DNN, and those who only have access to the output of the attacked model (label or confidence score), known as \textit{black-box attacks}. In addition, according to the objective to be reached, adversary attacks can also be distinguished as \textit{targeted} and \textit{untargeted} attacks. Formally, given an original input image $x$ and a trained classifier $C$, generating an adversarial example $x'$ can be formulated as a constrained optimization problem \cite{B9}:
\begin{align}
\label{eq1}
x'= \ \ &\underset{x'}{\arg\min} \ \mathcal{D}(x,x'), \nonumber\\ 
s.t. \ \ &C(x)=l, \nonumber\\ 
&C(x')=l',\\
&l \neq l,'\nonumber
\end{align}
where $\mathcal{D}$ denotes a distance metric between two data sample, while $l$ and $l'$ denote the output class label of $x$ and $x'$, respectively. In the case of a target attack, $l'$ is specified by the attacker, while for an untargeted attack, $l'$ can be any class label, as long as it is different from the correct label  $l$.

The distance metric $\mathcal{D}$ is used to quantify similarity/fidelity between the adversarial example and the original image. In the literature, three metrics are commonly used for generating adversarial examples, and all three are $L_{p}$ norms \cite{B10}. In other words, the amount of perturbation is quantified by $L_{p}$ norms, \textit{i.e.,} $\left\| x-x' \right\|_{p}$,  where the $p$-norm is defined as
\begin{equation}
\label{eq2}
\left\| v \right\|_{p}= \left( \sum^{n}_{i=1} |v_{i}|^{p} \right)^{\frac{1}{p}}
\end{equation}
Specifically, $L_{0}$, $L_{2}$ and $L_{\infty}$ are the three widely used metrics:
\begin{enumerate}
	\item $L_{0}$ metric counts the number of pixels that have been altered in the adversarial example.
	\item $L_{2}$ metric measures the Euclidean distance between the adversarial example and the original image.
	\item $L_{\infty}$ metric denotes the largest absolute difference value among all pixels in the adversarial example.
\end{enumerate}
\begin{table}[t!]
\centering
\caption{Selected adversarial attacks parameters with their corresponding values.}
\label{tab1}
\begin{tabular}{|l|c|c|}
\hline
Adversarial attack  & Parameter         & Values            \\ \hline
FGSM \cite{B5}     & $\epsilon$                        & 0.002, 0.03, 0.06, 0.14, 0.4                                               \\ \hline
BIM \cite{B16}      & $\epsilon$                        & 0.003, 0.03, 0.06, 0.15, 3                                                    \\ \hline
Deepfool \cite{B18} & overshoot                   & 0.25, 1.0, 3.5, 36, 500                                                              \\ \hline
C\&W     \cite{B10} & \begin{tabular}[c]{@{}c@{}}(confidence,\\ learning\_rate)\end{tabular} &  \begin{tabular}[c]{@{}c@{}} (10, 0.4), (10, 1), \\(30, 0.9), (30, 1.3), (70, 0.9)\end{tabular} \\ \hline
PGD     \cite{B20} & $\epsilon$                        & 0.003, 0.03, 0.1, 0.4, 1.40                                                 \\ \hline
MIM \cite{B19} & $\epsilon$                        & 0.005, 0.03, 0.06, 0.19, 0.6                                                     \\ \hline

\end{tabular}
\end{table}

Nevertheless, these metrics do not correlate with human perception, because, they totally overlook spatial relationships between image pixels, and also consider that all changes in the visual signal are of equal importance. Finally, they do not take into account any of the perceptual properties of the HVS. 

Several methods have been proposed in the literature to generate adversarial examples, they differ mainly in the modeling of objective function that seeks the best solution to the optimization problem described above in (\ref{eq1}). The perturbation is determined by maximizing the classification error, while minimizing distance metric. 

In our subjective study, we considered six prominent attacks that are: Fast Gradient Sign Method (FGSM) \cite{B5}, Basic Iterative Method (BIM) \cite{B16}, Deepfool \cite{B18}, Carlini-Wagner (C\&W) attack \cite{B10}, Projected Gradient Descent (PGD) \cite{B20} and Momentum Iterative Method (MIM) \cite{B19}. All these attacks are gradient-based adversarial generating approaches \cite{B9}. Specifically, the input image is perturbed according to the gradient of the loss function of the attacked DNN, where the perturbation magnitude gradually increases until the image is misclassified. For a complete description of these attacks, the reader is refereed to their original papers.
\section{Subjective Evaluation}
\label{sec3}
In this section, the conducted subjective study of adversarial examples is presented. Our goal is to use the ground truth obtained from human judgments to check the suitability of several state-of-the-art image fidelity assessment (IFA) metrics for adversarial examples, which can constitute viable alternatives to the three widely used distance metrics ($L_{0}$, $L_{2}$ and $L_{\infty}$). 
\subsection{Adversarial Attacks Description}
\begin{figure}[t!]
\centering
\subfigure{\label{S2a}\includegraphics[scale=0.27]{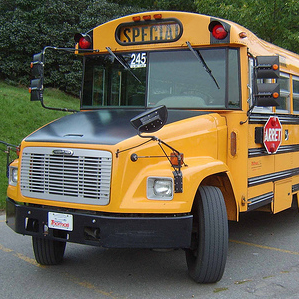}}
\subfigure{\label{S2b}\includegraphics[scale=0.27]{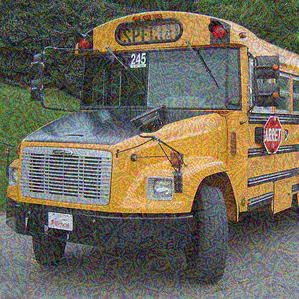}}
\subfigure{\label{S2c}\includegraphics[scale=0.27]{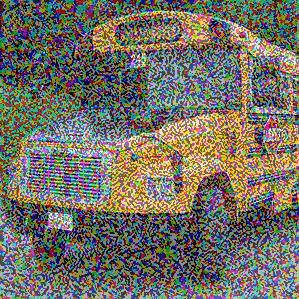}}\addtocounter{subfigure}{-3}\\
\subfigure[Small impairment]{\label{S2e}\includegraphics[scale=0.27]{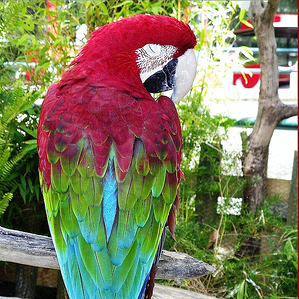}}
\subfigure[Medium impairment]{\label{S2f}\includegraphics[scale=0.27]{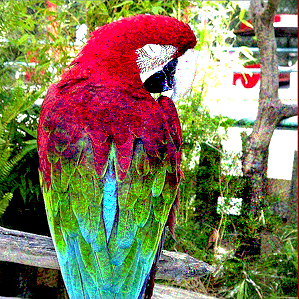}}
\subfigure[High impairment]{\label{S2g}\includegraphics[scale=0.27]{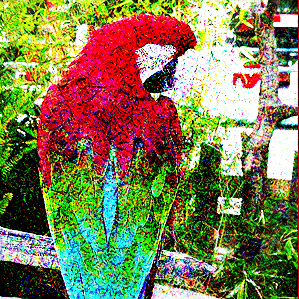}}
\caption{Adversarial examples with different levels of impairment generated using BIM (top) and C\&W (bottom) attacks.}
  \label{S2}
\end{figure}

As mentioned previously, a total of six adversarial attacks have been employed to generate the adversarial examples. All these attacks have been implemented using \textit{Cleverhans} software library \cite{B21}, which provides standardized reference implementations of adversarial example generation techniques. Each attack can be tuned through a set of parameters, here, we are only focused on the ones controlling the magnitude of the perturbation introduced. Table \ref{tab1} lists the parameters used to generate the different adversarial examples. The parameters values have been carefully chosen in a way to generate adversarial examples with a broad range of perturbations/distortions, thus covering the full range of subjective impairment scale, from imperceptible levels to high levels of impairment. \hbox{Figure \ref{S2}} shows some samples of adversarial images, in addition, the histogram of subjective scores for the entire dataset is illustrated in Figure \ref{S3}. 

As a victim DNN model, we used the well-known \hbox{Inception v3} network \cite{B22}, because it is pre-trained on \textit{ImageNet} dataset \cite{B23} that we considered as a source image, as reported in Section \ref{sec3.2}. Thus, the gradient of its loss function is exploited by the different attacks to compute the perturbation introduced to the original input image.   
\subsection{Dataset Preparation}
\label{sec3.2}
\begin{figure}[t!]
\centering
\includegraphics[scale=0.5]{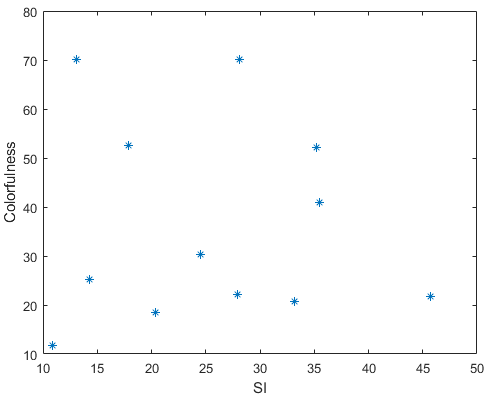}
\caption{SI and CF distributions of the selected contents.}
\label{S4}
\end{figure}

Since our work deals with adversarial examples for DNN-based image classification, we focused our subjective experiment on the most widely used image classification dataset, which is \textit{ImageNet} database \cite{B23}. Twelve images have been selected from the database that represent different content, including indoor and outdoor scenes and a wide range of colors and textures. In order to cover a wide range of features, the spatial complexity and color features of each image have been analyzed using Spatial Information (SI) \cite{B24} and ColorFulness (CF) \cite{B25}, respectively. The Figure \ref{S4} shows the values of SI and CF for all the selected images.

The original images have different sizes, that we cropped to the size of $299 \times 299$ pixels covering the main object in the image. Because, given that we used Inception v3 network as attacked DNN, and the latter has an image input size of $299 \times 299$. Consequently, to avoid the up- and down- sampling operations that can introduce distortions to the input image, we made choice to crop the images to the input size of the Inception v3 network.

Thus, the twelve selected and cropped images were used to produce the subjective test dataset. Each image was perturbed/attacked using the six different adversarial attacks with the five different parameter settings, thus providing 360 adversarial examples. In addition, two other different images have been selected for training.  
\subsection{Environment Setup and Test Methodology}
The subjective evaluations were conducted in a laboratory psychovisual test room, calibrated according to ITU-R BT.500-13 Recommendations \cite{B26}, equipped with a controlled lighting system and the color of the all background walls and curtains is mid-gray. A full HD 27-inch Dell UltraSharp U2717D was used to display the test stimuli. The distance of the subjects from the monitor was approximately equal to 7 times the picture height, as recommended in \cite{B27}. 

Since the detection of impairment is an important factor in our study, the subjective experiments have been conducted using the Double Stimulus Impairment Scale (DSIS) \hbox{method \cite{B26}}. Both the original image and adversarial example were displayed in a side-by-side arrangement on the same monitor. The original image and adversarial example were always displayed on the left and right side, respectively, and the subjects were aware of these positions.

At the end of the presentation of each pair of images, a dedicated user interface was displayed on the screen for about five seconds during which the subject gives its judgment. The participants were asked to rate the level of impairment of the adversarial examples with respect to the reference original image, using a five-grade discrete impairment scale (1: very annoying, 2: annoying, 3: slightly annoying, 4: perceptible, but not annoying, 5: imperceptible). In other words, the observers tried to quantity the visibility degree of the perturbation introduced by the attack.

Given the large number of stimuli, making impossible to show all of them in a single session, because the viewing session would exceed 30 minutes. Consequently, in order to avoid visual fatigue effects, the subjective experiment was divided into three sessions whose duration does not exceed 20 minutes each. Subjects took a break between each two sessions. Moreover, each test session involved only one subject assessing the stimuli. In order to avoid possible contextual and memory effects, the display order of these stimuli was randomized in a way that the same content was never shown consecutively.

Before the experiment starts, instructions explaining the task were provided to subjects. In addition, training session was held with additional images, allowing the subjects to practice and become familiarize with the test procedure. The quality of these training samples was chosen so that it covers the full rating scale.

A total of 18 naive subjects (5 females and 13 males) took part in the subjective experiment. The age of subjects was ranging from 21 to 54, with an average of 28.8. All subjects were screened for color blindness and visual acuity using Ishihara and Snellen charts, respectively.
\subsection{Data Processing}
First, the subjective scores were processed to detect and exclude possible outliers, \textit{i.e.,} subjects whose scores deviated strongly from others. Outliers detection was performed as specified in \cite{B26}, and no outlier subjects were detected in this study.

Second, the Mean Opinion Score (MOS) was computed as the mean across scores provided by different subjects as follows:
\begin{equation}
\label{eq3}
MOS_{j}=\frac{1}{N} \sum^{N}_{i=1}s_{ij}
\end{equation}
where $N$ is the number of subjects and $s_{ij}$ is the score given by subject $i$ for the stimulus $j$.

In order to evaluate the reliability of the obtained results from statistical point of view, 95\% confidence intervals (CI), assuming a Students \textit{t}-distribution of the scores, were computed together with MOS values.
\begin{figure}[t!]
\centering
\includegraphics[scale=0.5]{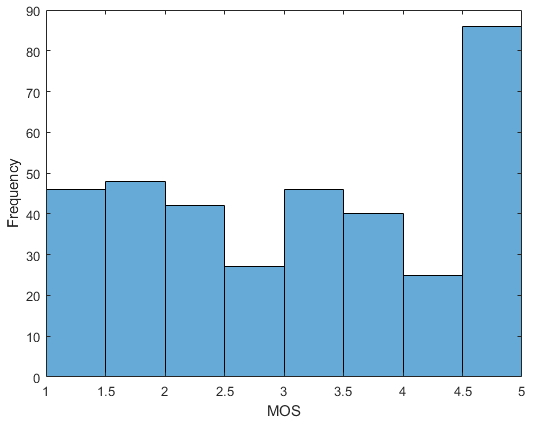}
\caption{Histogram of the MOS scores in the database.}
\label{S3}
\end{figure}
\section{Objective Evaluation and Results Analysis}
\label{sec4}
\begin{figure}[t!]
\centering
\includegraphics[scale=0.55]{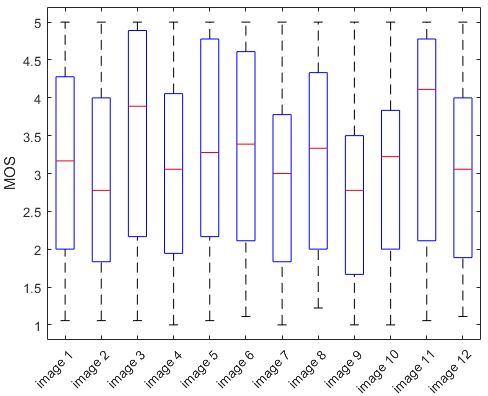}
\caption{The distribution of MOS values for each image.}
\label{S10}
\end{figure}

It is highly desirable that the obtained MOS scores show fair distribution of values and are representative of the different impairment level on the rating scale. Figure \ref{S3} shows MOS values distribution on the whole database. Overall, we obtained an almost a fair distribution, except for the 4.5-5 scale for which we obtained higher frequency. This mainly due to Deepfool attack, which impairment level is hardly to adjust and often provides undetectable perturbations. In addition, Figure \ref{S10} illustrates the distribution of MOS values for each assessed image. Thus, the resulting MOS values uniformly span the whole impairment scale, which means that the subjective experiments have been properly designed and conducted.

The results of the subjective tests were used as ground truth to evaluate fifteen full reference (FR) objective fidelity/quality metrics, namely: Peak-Signal-to-Noise-Ratio (PSNR), Structural Similarity Index (SSIM) \cite{B28}, Feature Similarity Index (FSIM/FSIMc for color images) \cite{B29}, Visual Signal-to-Noise Ratio (VSNR) \cite{B30}, Gradient Similarity Measure (GSIM) \cite{B31},  Most Apparent Distortion (MAD) \cite{B32}, Multi-Scale SSIM index (MS-SSIM) \cite{B33}, Visual Saliency-based Index (VSI) \cite{B34}, Visual Information Fidelity (VIF/VIFp for pixel domain) \cite{B35}, Information Fidelity Criterion (IFC) \cite{B36}, Weighted Signal-to-Noise Ratio (WSNR), Universal Quality Index (UQI) \cite{B37}, Noise Quality Measure (NQM) \cite{B38}.

In addition, the three widely used distance metrics ($L_{0}$, $L_{2}$ and $L_{\infty}$) have been also considered for evaluation and are compared against the fifteen objective fidelity metrics.
\begin{table}[t!]
\centering
\caption{Performance comparison of objective fidelity/quality metrics and $L_{p}$ norms distance metrics.}
\label{tab2}
\begin{tabular}{|l|c|c|c|c|}
\hline
\textbf{Method} & \multicolumn{1}{c|}{\textbf{PLCC}} & \multicolumn{1}{c|}{\textbf{SROCC}} & \multicolumn{1}{c|}{\textbf{RMSE}} & \multicolumn{1}{c|}{\textbf{OR}} \\ \hline
SSIM            & 0.936                         & 0.939                         & 0.416                        & 0.152                       \\ \hline
MS-SSIM         & 0.858                        & 0.942                         & 0.677                           & 0.152                       \\ \hline
VSI             & 0.876                         & 0.955                           & 0.634                         & 0.138                       \\ \hline
VIF             & 0.925                         & 0.932                          & 0.500                         & 0.172                       \\ \hline
VIFp            & 0.913                         & 0.925                          & 0.536                         & 0.172                       \\ \hline
MAD             & \textbf{0.977}                        & \textbf{0.973}                         & \textbf{0.275}                         & \textbf{0.119}                       \\ \hline
WSNR            & 0.941                         & 0.936                          & 0.445                         & 0.158                       \\ \hline
FSIM            & 0.891                         & 0.943                          & 0.598                         & 0.166                      \\ \hline
FSIMc           & 0.900                         & 0.944                          & 0.574                         & 0.163                      \\ \hline
PSNR            & 0.962                         & 0.958                          & 0.357                        & 0.138                      \\ \hline
UQI             & 0.901                         & 0.907                         & 0.571                         & 0.186                       \\ \hline
IFC             & 0.922                        & 0.914                         & 0.509                         & 0.166                       \\ \hline
NQM             & 0.942                         & 0.936                          & 0.441                         & 0.172                       \\ \hline
GSIM            & 0.835                        & 0.954                          & 0.725                         & 0.147                       \\ \hline
VSNR            & 0.923                        & 0.917                         & 0.507                         & 0.186                      \\ \hline
\hline
$L_{0}$         & 0.885                         & 0.915                         & 0.613                         & 0.186                       \\ \hline
$L_{2}$         & 0.914                         & 0.958                         & 0.533                         & 0.138                       \\ \hline
$L_{\infty}$    & 0.517                         & 0.645                         & 1.12                        & 0.336                       \\ \hline
\end{tabular}
\end{table} 

The performance evaluation of the set of metrics has been carried out in terms of three attributes: accuracy, monotonicity, and consistency, with respect to subjective scores. To achieve this goal, four performance measures were used, namely Pearson Linear Correlation Coefficient (PLCC) and Root Mean Square Error (RMSE) for prediction accuracy, while Spearman Rank Order Correlation Coefficient (SROCC) and Outlier Ratio (OR) for monotonicity and consistency, respectively.  We can say that a metric obtains good performance, if the values of PLCC and SROCC are high (close to $\pm 1$), and the values of RMSE and OR is low (close to 0).

PLCC measure was computed between the MOS and the objective score ($Q_{p}$) provided by the metric after a non-linear regression. This regression is performed using a 5-parameter logistic function as recommended in \cite{B14} and defined as follows:
\begin{equation}
\label{eq4}
Q_{p}(x)=\beta_{1}(\frac{1}{2}-\frac{1}{1+\exp^{\beta_{2}(x-\beta_{3})}})+\beta_{4}x+\beta_{5}
\end{equation}
where $\beta_{i} \ (i \in \{1, 2, 3, 4, 5\})$ are five free-parameters to be fitted based on the Gauss-Newton method.

The PLCC, SROCC, RMSE and OR results are provided in Table \ref{tab2}, where the top performing metric is given in boldface. Overall, a little more than half of the evaluated FR objective metrics provide good performance, especially MAD metric that shows the highest correlation with subjective scores. 

As expected, the $L_{p}$ distance metrics provide poor performance, except the $L_{2}$ distance that can be considered as acceptable, but still below those provided by the FR objective metrics. For instance, $L_{\infty}$ distance has obtained the worst results compared to all evaluated metrics.

According to the reporting results, most of the FR objective metrics provide better performance than the widely used $L_{p}$ distance metrics. Thanks to the inclusion of HVS features, the evaluated objective metrics correlate well with subjective scores and represent an obvious alternative to the $L_{p}$ distance metrics. Consequently, the adoption and inclusion of FR objective metrics in the construction of adversarial attacks can produce more optimal results, thus allowing to contribute in developing more robust deep neural networks.
\section{Conclusions}
\label{sec5}
In this paper, we focused on the visual fidelity assessment of adversarial examples. We presented a publicly available dataset of adversarial examples, which can be used to the design and evaluation of new objective IFA metrics specifically developed for this kind of impairment. The dataset was constructed through subjective experiment, where the original images as well as adversarial examples, along with objective and subjective scores are provided.

The test results clearly exhibited that the $L_{p}$ norms are non-suitable to quantify the perceived perturbations of adversarial examples, and that the objective fidelity/quality metrics represent a solid alternative to be a substitute for $L_{p}$ norms.

\end{document}